\documentclass[runningheads]{llncs}

\usepackage{eccv}

\usepackage{eccvabbrv}

\usepackage{graphicx}
\usepackage{booktabs}

\usepackage[accsupp]{axessibility}  

\usepackage{hyperref}

\usepackage{orcidlink}

\usepackage{pifont}
\definecolor{my_green}{RGB}{51,102,0}
\definecolor{my_red}{RGB}{204, 0, 0}
\newcommand{\cmark}{\textcolor{my_green}{\ding{51}}} 
\newcommand{\xmark}{\textcolor{my_red}{\ding{55}}} 

\usepackage{multirow}
\usepackage{graphicx}
\usepackage{tcolorbox}

\begin{document}

\title{{VITATECS}: A Diagnostic Dataset for Temporal Concept Understanding of Video-Language Models} 

\titlerunning{{VITATECS}: A Benchmark for Temporal Understanding Evaluation}


\author{Shicheng Li\inst{1} \and
Lei Li\inst{2} \and
Yi Liu\inst{1} \and
Shuhuai Ren\inst{1} \and
Yuanxin Liu\inst{1} \\
Rundong Gao\inst{1} \and
Xu Sun\inst{1} \and
Lu Hou\inst{3}
}

\authorrunning{S. Li et al.}

\institute{State Key Laboratory of Multimedia Information Processing, School of Computer Science, Peking University \\
\email{\{lisc99, xusun\}@pku.edu.cn} \and
The University of Hong Kong  \and
Huawei Noah’s Ark Lab \\
\email{houlu3@huawei.com}
}

\maketitle

\begin{abstract}

The ability to perceive how objects change over time is a crucial ingredient in human intelligence. 
However, current benchmarks cannot faithfully reflect the temporal understanding abilities of video-language models (VidLMs) due to the existence of static visual shortcuts. 
To remedy this issue, we present \textbf{VITATECS}, a diagnostic \textbf{VI}deo-\textbf{T}ext d\textbf{A}taset for the evaluation of \textbf{TE}mporal \textbf{C}oncept under\textbf{S}tanding. 
Specifically, we first introduce a fine-grained taxonomy of temporal concepts in natural language in order to diagnose the capability of VidLMs to comprehend different temporal aspects. 
Furthermore, to disentangle the correlation between static and temporal information, we generate counterfactual video descriptions that differ from the original one only in the specified temporal aspect. We employ a semi-automatic data collection framework using large language models and human-in-the-loop annotation to obtain high-quality counterfactual descriptions efficiently. Evaluation of representative video-language understanding models confirms their deficiency in temporal understanding, revealing the need for greater emphasis on the temporal elements in video-language research. 
Our dataset is publicly available at \url{https://github.com/lscpku/VITATECS}.

\end{abstract}

\section{Introduction}
\label{sec:intro}

\begin{figure}[tb]
  \centering
  \includegraphics[width=0.5\linewidth]{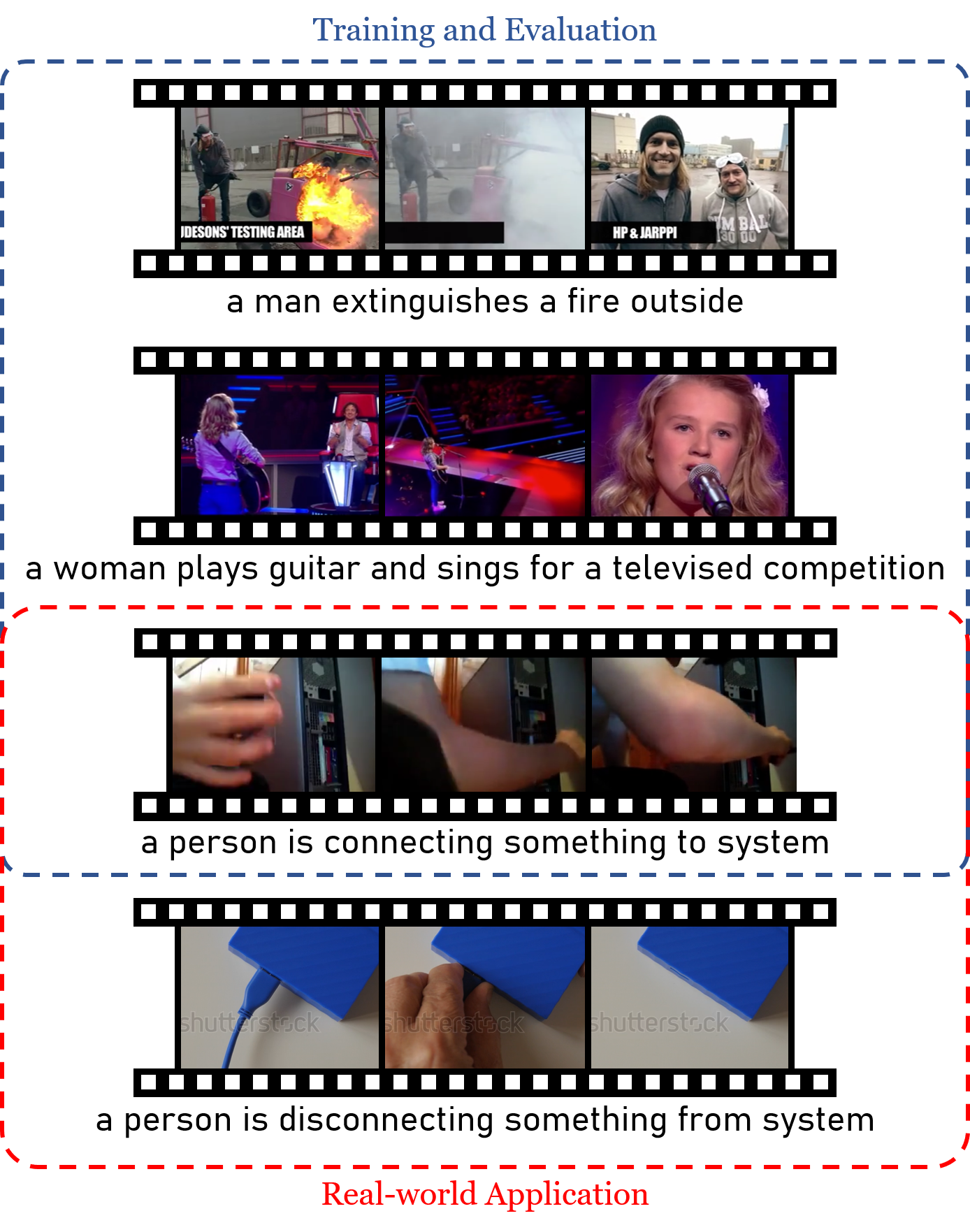}
  \caption{Illustration of the gap between current training and evaluation procedures and real-world applications. In current video-language datasets (\textcolor{blue}{blue} box), temporal information is highly correlated with static scenes. Models trained and evaluated on them cannot acquire the ability to understand temporal concepts, leading to failure in challenging real-world applications (\textcolor{red}{red} box).}
  \label{fig:gap}
\end{figure}

Many important concepts in human languages contain a temporal dimension~\cite{evans200422,klein2013time}, such as human actions, changes in status, and event order, which are beyond the expressive power of individual static images.
Such temporal concepts bring great challenges to video-language learning and are crucial for the generalization capability of intelligent systems in real-life scenarios. 

Although these temporal concepts are present in existing text-to-video retrieval~\cite{Xu2016MSRVTTAL,Wang2019VaTeXAL,Hendricks2018LocalizingMI} or video question answering~\cite{Xiao2021NExTQANP,Yu2019ActivityNetQAAD} benchmarks, most of these datasets fail to faithfully assess the temporal understanding ability of Video-Language Models (VidLMs) due to the strong correlation between static objects/scenes and temporal information. 
For example, in the blue box in \cref{fig:gap}, 
each video can be aligned to its description by merely identifying the static objects such as the fire, the microphone, and the PC case. 
As a consequence, the models may learn to simply rely on static clues to make predictions, leading to failure in real-world applications that require a genuine understanding of temporal concepts, \eg, to distinguish between the action of ``connecting something to system'' and ``disconnecting something from system'' as demonstrated by the red box in \cref{fig:gap}. 
Previous works~\cite{Huang2018WhatMA,Buch2022RevisitingT,SevillaLara2019OnlyTC,Lei2022RevealingSF,Girdhar2019CATERAD} have pointed out similar issues and provided several solutions. However, they do not properly define and categorize different aspects of temporal information. 
The lack of a clear definition adds to the difficulty of assessing the precise abilities of VidLMs. 
Additionally, they often construct evaluation datasets by following certain templates or using synthetic scenes, making them unsuitable for more diverse and realistic scenarios. 

In light of the drawbacks of current video-language testbeds, we propose a new dataset for VidLMs, \textbf{VITATECS}, to fill the gap for temporal concept understanding evaluation by decoupling temporal information and static information.
Inspired by Winoground~\cite{Thrush2022WinogroundPV}, to measure the ability of VidLMs to understand and align the temporal concepts, we ask the models to distinguish between the correct caption of a video and a modified version of the caption which contains similar static information and only differs in temporal information. 
To allow for a more comprehensive and fine-grained evaluation of temporal understanding ability, we summarize several aspects of temporal concepts that are commonly present in video descriptions, including \emph{Direction}, \emph{Intensity}, \emph{Sequence}, \emph{Localization}, \emph{Compositionality} and \emph{Type}, which according to our study cover most of the temporal information in video-language datasets. 

Since collecting high-quality video-text pairs is time-consuming and expensive, we follow previous works in dataset construction~\cite{Liu2022WANLIWA,Schick2021GeneratingDW,Park2022ExposingTL,Ye2022ProGenPZ,Meng2022GeneratingTD}, and augment existing open-domain video-language datasets by harnessing the world knowledge encoded in pre-trained large language models (LLMs)~\cite{chatgpt}. 
Specifically, given an annotated video-text pair in the dataset, we ask the LLM to generate a counterfactual description that only differs from the original description in one given temporal aspect using in-context learning~\cite{gpt3}.
To prevent potential mismatch when dealing with complex instructions, we design a human-in-the-loop procedure to filter out low-quality generations by iteratively generating counterfactual descriptions, human labeling, and fine-tuning a filter model.
In each iteration, the generated samples are used to update the filter model and the in-context learning exemplar set to boost generation and filtering quality. 
This annotation framework allows us to construct a 13k+ dataset from 231 human-written counterfactuals while maintaining high quality and diversity. 

Based on our dataset, we conduct a comprehensive evaluation of state-of-the-art video-language understanding models. 
Our findings can be summarized as follows. 
\begin{itemize}
    \item Existing models barely surpass random guesses in many aspects,  confirming their general lack of temporal understanding.
    \item Temporally-adapted image-text models outperform video-text pre-training, but primarily due to better utilization of static clues. 
    \item Failure of text encoders to learn temporal concepts during pre-training is partly responsible for low performance on temporal understanding. 
    \item Different video-text datasets tend to invoke different temporal understanding abilities. 
\end{itemize}
In summary, our work with VITATECS sheds light on limitations in current VidLMs' temporal understanding, providing insights for future development.

\section{Related Work}

\begin{table}[tb]
    \centering
    \caption{Comparison with other diagnostic video datasets from four aspects: whether they are video-language datasets, whether they are open-domain, whether they target temporal understanding ability, and whether they contain a fine-grained evaluation of model abilities. }
    \begin{tabular}{@{}lcccc@{}}
         \toprule
     Dataset    & video-language & open-domain & temporal & fine-grained \\
         \midrule
         Temporal Dataset~\cite{SevillaLara2019OnlyTC} & \xmark & \cmark & \cmark & \xmark \\
         CATER~\cite{Girdhar2019CATERAD} & \xmark & \xmark & \cmark & \xmark \\
         CLEVRER~\cite{Yi2019CLEVRERCE} & \cmark & \xmark & \cmark & \cmark \\
         SSv2-label~\cite{Lei2022RevealingSF} & \cmark & \xmark & \cmark & \xmark \\
         Contrast set~\cite{Park2022ExposingTL} & \cmark & \cmark & \xmark & \xmark \\
         VITATECS (Ours) & \cmark & \cmark & \cmark & \cmark \\
         \bottomrule
    \end{tabular}
    \label{tab:dataset_comparison}
\end{table}

\paragraph{Video-Language Understanding.}
With the great success of end-to-end deep learning models in natural language processing and image-text understanding, the research community has shown a growing interest in the more challenging task of video-language understanding, with promising results achieved on a wide range of tasks including video captioning~\cite{Krishna2017DenseCaptioningEI, Zhang2020DCADC}, video question answering~\cite{Xiao2021NExTQANP,Yu2019ActivityNetQAAD} and video-text retrieval~\cite{Xu2016MSRVTTAL,Hendricks2018LocalizingMI,Wang2019VaTeXAL,Chen2011MSVD,Rohrbach2015LSMDC,Zhou2017YouCook}. 
Some research~\cite{Ren2023TESTA,li2021alpro,fu2021violet,Fu2022violetv2,lei2021clipbert,cheng2022vindlu,Wang2022OmniVLOF,Wang2022AllIO,Li2020HeroHE,Luo2020UniViLMAU,Li2022LAVENDERUV} follows the prevalent paradigm in NLP and multi-modal understanding by directly conducting pre-training on video-text pairs. 
Another line of work~\cite{Luo2021CLIP4ClipAE,Ma2022XCLIPEM,Gorti2022XPoolCL,Bain2022CLIPHitchhiker,Gao2021CLIP2TVAE,Wang2022DisentangledRL,Xue2022CLIPViPAP} adapts powerful image-text pre-trained models like CLIP~\cite{Radford2021CLIP} to transfer their knowledge to the video-language domain. 
Recent studies~\cite{Li2023VideoChatCV,Zhang2023VideoLLaMAAI,videochatgpt,timechat} have also explored the possibility of integrating LLMs with vision encoders to perform video-language understanding tasks. 
Despite these valuable efforts, we argue that the apparent prosperity of video-language understanding models still rests upon the power of image-language models and that more attention should be paid to their temporal understanding abilities. 

\paragraph{Datasets on Temporal Understanding.}
Although the temporal dimension is the primary difference between videos and images, it has not received proper acknowledgment from current model design and dataset construction processes in the video-language community. 
Previous works~\cite{Buch2022RevisitingT,SevillaLara2019OnlyTC,Ghodrati2018VideoTP,Bagad2023TestOT,Huang2018WhatMA,Lei2022RevealingSF,Wang2018PullingAO,Huang2021SelfsupervisedVR,Wang2020RemovingTB} have pointed out the lack of emphasis on temporal understanding abilities. 
Evidence of this negligence includes the insensitivity of models to the frame order of input videos~\cite{Yun2022TimeIM,SevillaLara2019OnlyTC}, several order-agnostic architecture designs with state-of-the-art retrieval performance~\cite{Luo2021CLIP4ClipAE,Gorti2022XPoolCL}, visualization of intermediate layer features or saliency maps~\cite{Huang2018WhatMA,Choi2019WhyCI}, and even the success of using single frames training to achieve promising results~\cite{Lei2022RevealingSF}. 
Although a few datasets have been proposed to address this issue, they either fail to properly define and categorize different aspects of temporal information in video-language understanding, or only focuses on certain narrow aspects of temporality~\cite{DBLP:conf/cvpr/WeiLZF18,retro-actions,speednet,star,fetv}. 
In addition, many of theses benchmarks use videos constructed from computer-rendered synthetic videos~\cite{Girdhar2019CATERAD,Yi2019CLEVRERCE} or videos focusing on single human actions~\cite{Goyal2017TheS,SevillaLara2019OnlyTC,finegym,diving48}, which are not representative of real-world videos. 
We remedy these problems by identifying aspects of temporality and introducing a new benchmark for measuring the temporal understanding abilities in VidLMs that boasts higher diversity in video content and language forms. See \cref{tab:dataset_comparison} for a comparison between VITATECS and some existing video datasets. 


\section{VITATECS: Diagnosing Temporal Concept Understanding}
\label{sec:method}
In this section, we propose VITATECS, a new dataset for measuring how well VidLMs capture temporal information across modalities. It consists of (video, caption, counterfactual) triples, where the counterfactual description retains the same static information as the original caption while modifying its temporal information in one of the six fine-grained aspects that we define in \cref{sec:aspect}. 
We elaborate on the details of our temporal dataset in \cref{subsec:dataset_format} and the human-in-the-loop annotation framework we devise to facilitate its construction process in \cref{subsec:human_in_the_loop}.

\subsection{Fine-Grained Temporal Understanding}
\label{sec:aspect}

\begin{figure}[tb]
    \centering
    \includegraphics[width=\linewidth]{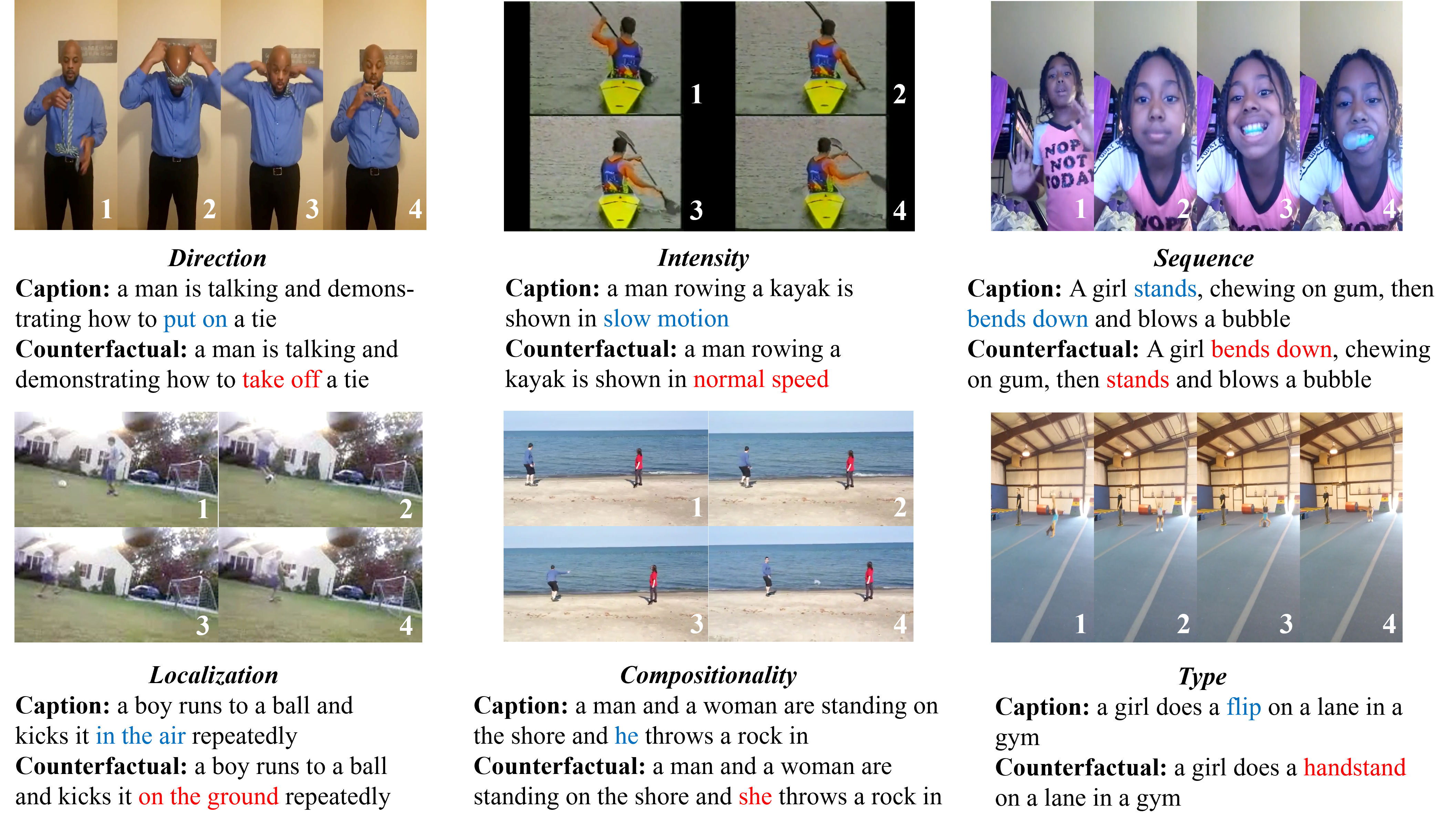}
    \caption{Examples from the six aspects of our dataset. Each sample contains a video, a ground-truth caption, and a counterfactual description with modifications in the given temporal aspect. Differences between the sentence pairs are highlighted in \textcolor{blue}{blue} and \textcolor{red}{red}.}
    \label{fig:examples} 
\end{figure}

Measuring the temporal understanding ability of VidLMs is a challenging task. 
On one hand, it is not clear how to define and characterize the temporal information in a video. 
Previous works~\cite{SevillaLara2019OnlyTC,Lei2022RevealingSF,Park2022ExposingTL} draw a rough equivalence between temporal information and the actions in the video. 
In reality, temporal information can emerge in a variety of forms, such as human actions, changes in object status, dynamics of substances, the order of events, \emph{etc.}, and is widely manifested in daily activities. 
On the other hand, it is infeasible to completely disentangle the temporal information from the static information. 
The background scenes, objects, and people's postures are all highly correlated with the temporal information in open-domain videos. 
If not properly controlled, such static bias would allow models to rely on static clues as shortcuts for making predictions while seemingly learning to capture the temporal information.

To achieve high coverage of temporal information in video-language datasets and allow for fine-grained diagnosis of temporal understanding abilities, we identify six aspects of temporal concepts commonly reflected in natural language: \emph{Direction}, \emph{Intensity}, \emph{Sequence}, \emph{Localization}, \emph{Compositionality} and \emph{Type}. 
These aspects of temporal information are disentangled from static information to different degrees and address different facets of the temporal information in video-language datasets, allowing us to pinpoint the temporal understanding abilities of VidLMs. 
Since our final target is to construct text pairs with aspect-specific modifications, for clarity, we define these aspects in terms of the temporal questions they address and the corresponding modification patterns as follows. 
\begin{itemize}
    \item \textbf{``Direction''} measures the model's ability to answer the following question: ``In which direction does the status of objects change?'' Examples of this aspect include sentence pairs describing opposite spatial movements or one action reversing the effect of the other. 
    \item \textbf{``Intensity'' } measures the model's ability to answer the following question: ``How fast or how intense does the change occur?'' Examples of this aspect include counterfactual sentences which change the words that modify the verbs or change the verb to a similar action with subtle differences in the manner it is conducted. 
    \item \textbf{``Sequence''} measures the model's ability to answer the following question: ``How many events are depicted in the video and in what order?'' Examples of this aspect usually involve changing the temporal order or number of occurrences of the events. 
    \item \textbf{``Localization''} measures the model's ability to answer the following question: ``On which part of the frame does the change occur?'' Examples of this aspect include sentence pairs with the same action conducted either in different absolute spatial locations or in different locations in relation to other objects in the video. 
    \item \textbf{``Compositionality''} measures the model's ability to answer the following question: ``Who performed which action and to whom?'' Examples of this aspect often include actions with interchanged subjects or objects. 
    \item \textbf{``Type''} measures the model's ability to answer the following question: ``What is the action depicted in the video?'' This aspect contains general alterations to the actions with a less stringent constraint on the static information contained. 
\end{itemize}

To validate the coverage of our temporal concept categorization, we randomly sample 200 video-text pairs from MSR-VTT~\cite{Xu2016MSRVTTAL} and VATEX~\cite{Wang2019VaTeXAL} and inspect the types of temporal information they contain. We find that for 98\% of the samples, their temporal information falls in one of our categories, which demonstrates that our taxonomy is able to achieve high coverage while taking into account the disentanglement from static information. 

\subsection{Dataset Format}
\label{subsec:dataset_format}
Following Winoground~\cite{Thrush2022WinogroundPV}, we measure the ability of VidLMs to match the videos to their correct descriptions among some well-designed choices as a proxy for their temporal understanding abilities. 
Specifically, for each aspect $a$, we collect (video, caption, counterfactual) triples $\{(V_i, C_i, \tilde{C}_i)\}_{i=1}^{N_a}$ where $N_a$ denotes the number of samples for aspect $a$, $V_i$ denotes the video, $C_i$ denotes the true caption of the video, and $\tilde{C}_i$ is the counterfactual description that differs from $C_i$ only in the temporal aspect $a$.  \cref{fig:examples} shows examples of our dataset.

\begin{figure}[tb]
    \centering
    \includegraphics[width=\linewidth]{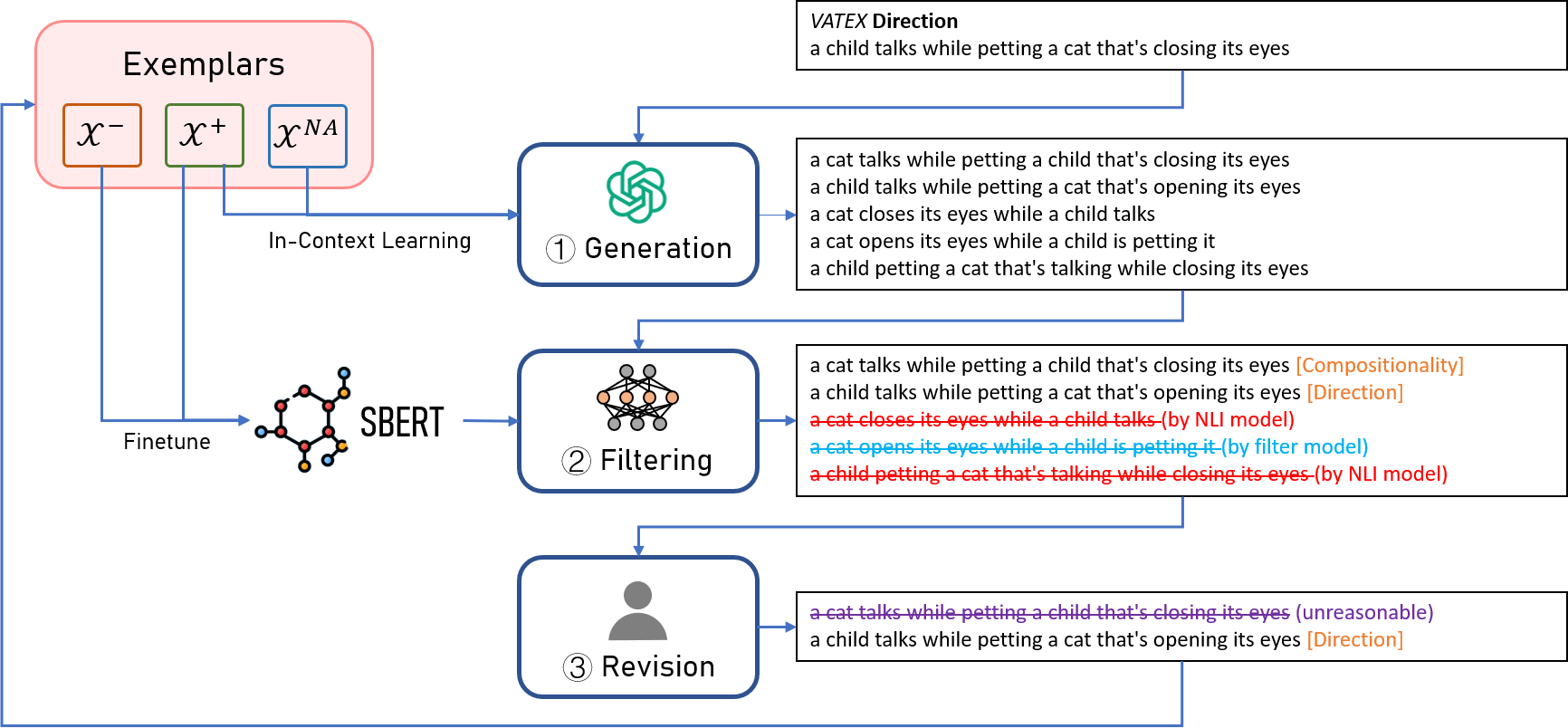}
    \caption{Illustration of our human-in-the-loop annotation framework. Texts in \textcolor{orange}{orange} indicate the labels predicted by the filter model. Texts in \textcolor{red}{red}, \textcolor{cyan}{blue} and \textcolor{violet}{purple} indicate candidates that are eliminated by the NLI model, the filter model and human annotators, respectively. }
    \label{fig:flowchart}
\end{figure}

\subsection{Human-in-the-Loop Annotation Framework}
\label{subsec:human_in_the_loop}

Due to the heavy expenses of collecting high-quality (video, caption, counterfactual) triples, we present a human-in-the-loop annotation framework for semi-automatic counterfactual generation based on existing (video, caption) datasets. 
At the core of our framework is a loop consisting of three stages: \textbf{generation}, \textbf{filtering}, and \textbf{revision}. 
In stage 1, we use in-context learning~\cite{gpt3,Dong2022ASF} to generate candidate counterfactuals based on ground-truth video-text pairs with LLMs. 
In stage 2, the candidates are filtered using a combination of rules, off-the-shelf language understanding models, and fine-tuned language understanding models. 
In stage 3, we ask human annotators to verify the quality of the candidates and use the high-quality ones to refine the generation process and the filter model. 
The three stages are conducted on a small subset and are repeated until the filter model achieves satisfactory precision on a held-out evaluation set. 
Below, we first lay out the criteria for our counterfactual descriptions and then elaborate on the details of each stage. 

\paragraph{Criteria.}

During our effort to construct the dataset, we found that LLMs encounter some difficulty in following our instructions when generating counterfactual descriptions, possibly due to the reflective nature of our temporal concepts. 
To enable consistent and high-quality counterfactual generation, we first identify five major criteria for measuring the quality of generated counterfactuals as follows. 
\begin{enumerate}
    \item[(a)] The counterfactual should neither entail nor be entailed by the caption. 
    \item[(b)] The counterfactual should contain roughly the same amount of information as the caption. 
    \item[(c)] The counterfactual should be grammatically correct and semantically plausible. 
    \item[(d)] The counterfactual should retain the static information in the caption and only change the given aspect of temporal information. 
    \item[(e)] The pattern of counterfactual description should be diverse across the entire dataset. 
\end{enumerate}

Among these desirable properties, criteria (a)-(d) are instance-level criteria we aim to address in both the generation and filtering stages. In contrast, criterion (e) is a dataset-level criterion dealt with in a finalization step after the filter model has converged. 

\paragraph{Exemplar Sets.}
Throughout our annotation process, we maintain three sets of exemplars: positive set $\mathcal{X}^+$ contains sentence pairs that differ only in a given aspect; negative set $\mathcal{X}^-$ contains sentence pairs that violate one of the aforementioned criteria (a)-(d); N/A set $\mathcal{X}^{NA}$ contains captions that do not describe a certain aspect of the temporal concept. 
These exemplars serve two purposes: on the one hand, they compose the demonstrations of valid and invalid data samples for in-context learning, which supply the generative language models with clearer and better-informed instructions; on the other hand, they provide supervision signals for the fine-tuning of the filter model. 
These three sets are initialized with manually annotated examples and expanded semi-automatically to boost the generation and filter model performance as more data samples are generated. 
The size and examples of the initial exemplar sets are available in the Appendix. 

\paragraph{In-Context Learning Generation.}

In this stage, we draw upon the generative strength of ChatGPT (\texttt{gpt-3.5-turbo-0613})~\cite{chatgpt} to generate counterfactual descriptions given the original caption and the desired aspect of variation. 
The use of in-context learning allows us to capture the different aspects of temporal concepts through carefully-designed instructions and demonstrations. 
Specifically, we first randomly sample a small subset (500 for each aspect) of (video, caption) pairs from the test sets of two popular video-text retrieval datasets, MSR-VTT~\cite{Xu2016MSRVTTAL} and VATEX~\cite{Wang2019VaTeXAL}. 
Then, for each (video, caption) pair, we invoke the instruction following and pattern replication abilities of ChatGPT by constructing a prompt consisting of an aspect-specific instruction, demonstrations sampled from the exemplar sets, and the query for which we aim to generate the counterfactual description. 
The demonstrations are sampled from both $\mathcal{X}^+$ and $\mathcal{X}^{NA}$ so that the LLM not only learns to generate counterfactual descriptions for valid captions but also learns to recognize which captions do not concern the temporal aspect of interest. 

\paragraph{Automatic Filtering.} 

In view of the uneven quality of generated examples, we propose to filter the candidates and automatize this procedure using natural language understanding models. 
First, we leverage an off-the-shelf natural language inference (NLI) model, Sentence-BERT~\cite{reimers-2019-sentence-bert}, to filter out examples that do not meet criterion (a), \emph{i.e.}, the cases where one description entails the other. 
Then, to filter out candidates that do not meet criterion (b)-(d), we use a neural network that takes a pair of sentences as input and performs a 7-way classification task, where category 0 corresponds to disqualified generations and categories 1-6 correspond to the six aspects we define. 
Considering the similarity in task formulation, we initialize the filter model with the same NLI model above. 
The fine-tuning data consists of samples from both $\mathcal{X}^+$ and $\mathcal{X}^-$.
We adopt a rigorous decision mechanism that classifies the given sentence pair into one of the six aspects only if the model makes consistent predictions for the pair and its reversed version with high confidence, as we care more about the precision of the filter model than its recall. 

\paragraph{Human Revision.}

To guarantee the quality of filtered examples and guide both the in-context learning procedure and the filter model in the right direction, we introduce human supervision to revise the filtering results. 
We manually check the samples that are predicted to fall in one of the six aspects and correct the wrong predictions. 
Note that, on the one hand, due to the relatively small size of the sampled subset and the rigorous confidence-based filtering procedure, the number of examples for human revision is reduced significantly; on the other hand, human annotators only need to rectify the predicted labels instead of writing the entire counterfactual description. 
Therefore, this revision stage does not require excessive human effort and only incurs acceptable annotation costs. 

\paragraph{Iterative Procedure.}
We repeat the generation, filtering, and revision procedure to iteratively enlarge the exemplar sets and refine the filter model. 
In each iteration, the previously revised examples are incorporated into $\mathcal{X}^+$ and $\mathcal{X}^-$ according to their labels. 
This simultaneously augments the demonstration set of in-context learning for better generation quality and provides more training data for fine-tuning the filter model.
After each iteration, the fine-tuned filter model is evaluated on an independently annotated test set. 
We terminate the iteration once no significant improvement of the filter model is observed. 

\paragraph{Finalization.}
After the filter model has converged, we perform generation and filtering on a larger scale (20,000 for each aspect) without human revision. 
As a finalization step, we address the issue of diversity by favoring generations that involve a less common change of verb throughout the dataset when merging the filtered samples. 

\begin{table}[tb]
  \caption{Statistics of our dataset including the number of samples, the number of videos, and the average length of the original captions and the counterfactual descriptions. }
  \centering
  \resizebox{\linewidth}{!}{
  \begin{tabular}{@{}lcccccc@{}}
    \toprule
         & Direction & Intensity & Sequence & Localization & Compositionality & Type \\
    \midrule
    \# samples & 3,800 & 779 & 151 & 1,053 & 1,450 & 6,605 \\
    \# videos  & 2,646 & 692 & 150 & 915 & 1,110 & 4,287 \\
    \midrule
    Avg. len (caption)     & 13.6 & 13.6 & 14.9 & 14.6 & 13.9 & 11.7 \\
    Avg. len (counterfactual)     & 13.8 & 13.9 & 14.9 & 14.5 & 13.9 & 11.6 \\
    \bottomrule
  \end{tabular}
  }
  \label{tab:stat}
\end{table}

\paragraph{Annotation Efficiency of the Framework.} Our framework can be easily scaled to generate larger datasets since no more human efforts are required once the filter model has converged. 
In our case, it only takes 231 human-written descriptions and around 1500 labeling annotations to obtain the final benchmark with 13k+ samples, showing the efficiency of our annotation framework. The statistics of our dataset are shown in \cref{tab:stat}. 
We also manually check the quality of VITATECS by sampling 100 instances from each aspect and find that 94.8\% of them satisfy our criteria. See Appendix A for more details on the quality check process.


\section{Evaluation of Video-Language Models}
\label{sec:exp}

In this section, we evaluate prevailing VidLMs to examine their temporal understanding ability.
We first introduce the evaluation settings and then discuss the findings drawn from our evaluation to facilitate future studies. 

\subsection{Experimental Setup}

\paragraph{Evaluated Models.} In our experiments, we focus on models designed for the video-text retrieval task, which can calculate the similarity score between a video and a text query. 
We test three pre-trained VidLMs (VIOLET~\cite{fu2021violet}, ALPRO~\cite{li2021alpro} and Singularity~\cite{Lei2022RevealingSF}) and three temporally-adapted image-language models (CLIP4Clip~\cite{Luo2021CLIP4ClipAE}, X-Pool~\cite{Gorti2022XPoolCL} and X-CLIP~\cite{Ma2022XCLIPEM}). 
We also include two recent video large language models, Video-LLaMA~\cite{Zhang2023VideoLLaMAAI} and VideoChat~\cite{Li2023VideoChatCV}, as well as pure image-text foundation models such as BLIP~\cite{Li2022BLIPBL}, which has shown strong performance on zero-shot video-text retrieval. 

\paragraph{Evaluation Metric.} A model's prediction is considered correct if the similarity score of the correct caption is higher than that of the generated counterfactual. 
We measure the accuracy of the models on each of the six aspects of temporal concepts, and explore a recall-based metric in \cref{subsec:discuss}.

\paragraph{Human Baseline.} We randomly choose 100 samples for each aspect from our dataset and ask five volunteers to help establish a human performance baseline. The annotators are shown a video and two text descriptions at a time and are required to choose the text that best describes the video. 
We report the average accuracy of the five annotators as the human baseline. 

\begin{table}[tb]
  \centering
  \caption{Accuracy (\%) of human annotators and state-of-the-art VidLMs on VITATECS. The VidLMs are evaluated on the full dataset, while human performance is marked in \textcolor{gray}{gray} to indicate it is evaluated only on a randomly sampled subset.}
  \resizebox{\linewidth}{!}{
  \begin{tabular}{@{}lcccccccc@{}}
    \toprule
     & Dataset &  Direction & Intensity & Sequence & Localization & Compositionality & Type & Avg. \\
    \midrule
    BLIP-large~\cite{Li2022BLIPBL} & Zero-shot & 58.6 & \textbf{67.7} & 51.7 & 66.2 & 61.8 & 78.6 & 64.1 \\
    Singularity~\cite{Lei2022RevealingSF} & MSR-VTT & 54.7 & 61.7 & 52.3 & 63.0 & \textbf{65.5} & 77.4 & 62.4 \\   
    ALPRO~\cite{li2021alpro} & MSR-VTT & 55.4 & 56.0 & 45.7 & 59.2 & 58.6 & 74.5 & 58.2 \\
    VIOLET~\cite{fu2021violet} & MSR-VTT & 60.2 & 62.8 & \textbf{61.6} & 60.6 & 64.8 & 78.2 & 64.7 \\
    CLIP4Clip~\cite{Luo2021CLIP4ClipAE} & MSR-VTT & 62.6 & 65.3 & 51.7 & \textbf{66.5} & 63.5 & 82.4 & \textbf{65.3} \\   
    X-Pool~\cite{Gorti2022XPoolCL} & MSR-VTT & 59.9 & 63.0 & 55.6 & \textbf{66.5} & 64.3 & 81.3 & 65.1 \\
    X-CLIP~\cite{Ma2022XCLIPEM} & MSR-VTT & \textbf{63.6} & 60.8 & 55.6 & 64.5 & 63.7 & \textbf{83.2} & 65.2 \\   
    Video-LLaMA~\cite{Zhang2023VideoLLaMAAI} & Zero-shot & 51.6 & 52.2 & 56.3 & 51.0 & 49.4 & 51.7 & 52.0 \\
    VideoChat~\cite{Li2023VideoChatCV} & Zero-shot & 52.3 & 50.3 & 46.4 & 50.4 & 51.7 & 51.0 & 50.4 \\
    \midrule
    \textcolor{gray}{Human} & - & \textcolor{gray}{94.6} & \textcolor{gray}{93.2} & \textcolor{gray}{94.0} & \textcolor{gray}{93.8} & \textcolor{gray}{97.8} & \textcolor{gray}{92.2} & \textcolor{gray}{94.3} \\
    \bottomrule
  \end{tabular}
  }
  \label{tab:results}
\end{table}

\subsection{Evaluation Results}

\paragraph{Overall Performance.} 
As shown in \cref{tab:results}, although humans can easily match the videos to their correct descriptions with high consistency ($\kappa=0.86$) and nearly no mistakes, the overall performance of all the evaluated models is still far from expectations. 
No model achieves an accuracy of over 70\% on the temporal aspects other than the relatively easy ``Type'' aspect, which has the strongest correlation with the static information. 
Particularly, on the more temporally demanding aspects (``Direction'', ``Intensity'', and ``Sequence''), the models perform barely over the random baseline (50\%). 
Considering that part of our videos directly comes from MSR-VTT, the poor performance of models fine-tuned on MSR-VTT reaffirms our statement that existing video-language datasets are incapable of assessing the temporal understanding ability of models. 

\begin{table}[tb]
  \centering
  \caption{Accuracy (\%) of CLIP-based models with and without temporal aggregation modules}
  \resizebox{\linewidth}{!}{
  \begin{tabular}{@{}lcccccccc@{}}
    \toprule
    Model & Temporal & Direction & Intensity & Sequence & Localization & Compositionality & Type & Avg. \\
    \midrule
    \multirow{2}{5.5em}{CLIP4Clip~\cite{Luo2021CLIP4ClipAE}} & \cmark & \textbf{62.6} & 65.3 & 51.7 & \textbf{66.5} & \textbf{63.5} & \textbf{82.4} & 65.3 \\
     & \xmark & 61.6 & \textbf{67.3} & \textbf{60.3} & 66.1 & 62.8 & \textbf{82.4} & \textbf{66.8} \\
    \midrule
    \multirow{2}{5.5em}{X-CLIP~\cite{Ma2022XCLIPEM}} & \cmark & \textbf{63.6} & 60.8 & 55.6 & 64.5 & 63.7 & \textbf{83.2} & 65.2 \\
     & \xmark & 62.1 & \textbf{63.8} & \textbf{59.6} & \textbf{65.6} & \textbf{64.2} & 82.6 & \textbf{66.3} \\
    \midrule
    \multirow{2}{5.5em}{X-Pool~\cite{Gorti2022XPoolCL}} & \cmark & \textbf{60.4} & \textbf{65.5} & \textbf{58.3} & 65.0 & 62.1 & 79.9 & \textbf{65.2} \\
     & \xmark & 59.9 & 63.0 & 55.6 & \textbf{66.5} & \textbf{64.3} & \textbf{81.3} & 65.1 \\
    \bottomrule
  \end{tabular}}
  \label{tab:res-temp}
\end{table}

\paragraph{Effects of Vision Encoders.} 
Among the models we evaluate, the temporally-adapted image-text models based on CLIP generally outperform the models with video-text pre-training. 
To further investigate how much the temporal aggregation modules contribute to the temporal understanding abilities of the CLIP-based models, we disable the temporal aggregation module in these models and replace it with a simple mean pooling layer. 
The results are shown in \cref{tab:res-temp}. 
Contrary to what is expected, disabling the temporal aggregation module only results in a slight drop in performance for X-Pool. It even improves the temporal understanding ability of CLIP4Clip and X-CLIP. 
This suggests that these temporal aggregation modules are potentially under-trained due to the weak requirement of temporal modeling in video-language datasets like MSR-VTT.
Consequently, the superiority of the CLIP-based models mainly stems from the effective utilization of the static information in the video instead of a true understanding of the temporal concepts. 
For a similar reason, image-text models are able to achieve comparable performance on our dataset without further video-text training. 

\begin{table}[tb]
  \centering
  \caption{Average cosine similarity between the representations of original captions and counterfactual descriptions produced by different text encoders}
  \resizebox{\linewidth}{!}{
  \begin{tabular}{@{}lccccccc@{}}
    \toprule
   Text Encoder & Direction & Intensity & Sequence & Localization & Compositionality & Type & Avg. \\
    \midrule
    CLIP-text~\cite{Radford2021CLIP} & 0.963 & 0.964 & 0.975 & 0.965 & 0.970 & 0.912 & 0.958 \\
    Sentence-BERT~\cite{reimers-2019-sentence-bert} & 0.890 & 0.940 & 0.970 & 0.916 & 0.939 & 0.704 & 0.893 \\
    CLIP4Clip~\cite{Luo2021CLIP4ClipAE} & 0.941 & 0.939 & 0.969 & 0.932 & 0.947 & 0.828 & 0.926 \\
    CLIP4Clip-temporal~\cite{Luo2021CLIP4ClipAE} & 0.946 & 0.946 & 0.971 & 0.939 & 0.953 & 0.847 & 0.934 \\
    \bottomrule
  \end{tabular}}
  \label{tab:res-text}
\end{table}

\paragraph{Similarity of Text Representations.} 
We calculate the average cosine similarity between the representations of the original captions and the counterfactual descriptions with different text encoders. 
As shown in \cref{tab:res-text}, both the CLIP text encoder and Sentence-BERT produce highly similar sentence representations for samples in the ``Sequence'' aspect, indicating that the struggle of the evaluated models can partly be explained by the inability of text encoders to recognize the temporal distinction between the captions and the counterfactual descriptions. 
We also notice that the CLIP text encoder generally produces higher similarity scores even after it is fine-tuned on video-text data. 
This suggests that the ability to identify temporal concepts in natural language may be lost during the image-text pre-training stage and cannot be recovered by fine-tuning on existing video-language datasets. 

\begin{table}[tb]
  \centering
  \caption{Accuracy (\%) of VIOLET, Singularity, and X-Pool fine-tuned on different video-text datasets}
  \resizebox{\linewidth}{!}{
  \begin{tabular}{@{}lcccccccc@{}}
    \toprule
    Model  & Dataset & Direction & Intensity & Sequence & Localization & Compositionality & Type & Avg. \\
    \midrule
    \multirow{3}{7em}{VIOLET~\cite{fu2021violet}}
    & DiDeMo~\cite{Hendricks2018LocalizingMI} & 50.9 & 59.7 & 55.6 & \textbf{61.6} & 64.5 & 77.7 & 61.7 \\
    & LSMDC~\cite{Rohrbach2015LSMDC} & \textbf{60.2} & \textbf{62.8} & 61.6 & 60.6 & \textbf{64.8} & \textbf{78.2} & \textbf{64.7} \\
    & YouCook2~\cite{Zhou2017YouCook} & 58.2 & 60.2 & \textbf{62.9} & 61.1 & 61.7 & 76.8 & 63.5 \\
    \midrule
    \multirow{3}{7em}{Singularity~\cite{Lei2022RevealingSF}} 
    & ActivityNet~\cite{Krishna2017DenseCaptioningEI} & 54.0 & 64.8 & 50.3 & 64.7 & 61.8 & 76.0 & 61.9 \\
    & DiDeMo~\cite{Hendricks2018LocalizingMI} & \textbf{57.1} & \textbf{65.3} & \textbf{53.6} & \textbf{67.2} & \textbf{64.3} & \textbf{76.9} & \textbf{64.1} \\
    & SSv2-label~\cite{Goyal2017TheS,Lei2022RevealingSF} & \textbf{57.1} & 65.1 & 49.7 & 63.5 & 59.4 & 75.2 & 61.7 \\
    \midrule
    \multirow{2}{7em}{X-Pool~\cite{Gorti2022XPoolCL}} 
    & LSMDC~\cite{Rohrbach2015LSMDC} & 60.1 & \textbf{69.2} & 50.3 & 66.6 & 59.4 & 77.1 & \textbf{63.8} \\
    & MSVD~\cite{Chen2011MSVD} & \textbf{64.4} & 57.9 & \textbf{51.0} & \textbf{68.3} & \textbf{62.1} & \textbf{78.8} & \textbf{63.8} \\
    \bottomrule
  \end{tabular}}
  \label{tab:res-data}
\end{table}

\paragraph{Effects of Fine-Tuning Data.} 
We conduct a comparison between the performance of VidLMs fine-tuned on different downstream datasets. 
The results are shown in \cref{tab:res-data}. 
We find that models fine-tuned on different text-to-video retrieval datasets exhibit different temporal understanding abilities. 
For example, DiDeMo tends to elicit higher accuracy on ``Localization'' and ``Compositionality'', while LSMDC contributes to better understanding of ``Intensity''.  
Also, since SSv2 only depicts single human actions, it brings benefits on the ``Direction'' aspect but not on ``Sequence'' understanding, which can be improved by fine-tuning on datasets with longer video duration and dense captions such as YouCook2. 
This finding advocates the use of diverse videos and captions in the training process. 

\subsection{Discussions}
\label{subsec:discuss}

\begin{table}[tb]
  \centering
  \caption{Recall@10 of ALPRO and CLIP4Clip on video-to-text retrieval on VITATECS}
  \resizebox{\linewidth}{!}{
  \begin{tabular}{@{}lcccccccccc@{}}
    \toprule
     Model & Dataset & Description & Direction & Intensity & Sequence & Localization & Compositionality & Type & Avg. \\
     \midrule
     \multirow{3}{5em}{ALPRO} & \multirow{3}{5em}{Zero-shot} & Caption & 28.5 & 48.1 & 70.3 & 43.9 & 34.6 & 22.9 & 41.4 \\
    & & Counterfactual & 28.2 & 42.9 & 70.0 & 38.4 & 32.6 & 12.7 & 37.5 \\
    & & All & 28.3 & 45.5 & 70.2 & 41.2 & 33.6 & 17.8 & 39.4 \\
    \midrule
    \multirow{3}{5em}{CLIP4Clip} & \multirow{3}{5em}{MSR-VTT} & Caption & 53.6 & 73.7 & 90.7 & 69.8 & 65.2 & 48.5 & 66.9 \\
    & & Counterfactual & 47.7 & 66.0 & 88.7 & 60.0 & 60.4 & 22.4 & 57.5 \\
    & & All & 50.7 & 69.9 & 89.7 & 64.9 & 62.8 & 35.4 & 62.2 \\
    \bottomrule
  \end{tabular}
  }
  \label{tab:recall10}
\end{table}

\paragraph{Recall on VITATECS. }

Previous work~\cite{Diwan2022WhyIW} on the challenges of Winoground points out that accuracies based on cosine similarity comparison might be too harsh for the models, and it is possible that they under-perform on Winoground because the image-text pairs are out-of-distribution for them. 
This is also a concern for our dataset, so we follow them by calculating the Recall at $k > 1$ on the task of video-to-text retrieval on the entire VITATECS dataset for each aspect. 
Since a video may have multiple caption-counterfactual pairs in our dataset, we choose $k = 10$ and show the recalls for captions, counterfactuals, and both descriptions in \cref{tab:recall10}. 
We observe that for both ALPRO and CLIP4Clip, the recalls of captions and counterfactuals are very close. 
This indicates that the models are able to connect the texts with their corresponding videos through the shared static information, but cannot distinguish between the different temporal information in the caption and the counterfactual.

\begin{table}[tb]
  \centering
  \caption{Accuracy (\%) of X-CLIP on VITATECS and other counterfactual construction strategies}
  \begin{tabular}{@{}lcccccccc@{}}
    \toprule
     POS of replaced words & \multicolumn{3}{c}{All} & Noun & Verb & Adjective \\
     \# replaced words & 1 & 2 & 3 & 1 & 1 & 1 \\
    \midrule
    Random & 74.5 & 82.7 & 90.3 & 82.1 & 67.0 & 71.3 \\
    Synonym & 64.8 & 77.2 & 83.3 & 72.1 & 64.8 & 67.5 \\
    \midrule
    VITATECS (subset) & \multicolumn{6}{c}{64.3} \\
    \bottomrule
  \end{tabular}
  \label{tab:ablation}
\end{table}

\paragraph{Ablation Study of Counterfactual Design.}
To verify the design of our counterfactual descriptions, we randomly sample 100 instances from each aspect of VITATECS and apply different modification strategies to the original captions. 
Specifically, we randomly choose 1-3 words in the caption and replace them with its synonym or a random word of the same part of speech. We also experiment with different types of words (nouns, verbs, or adjectives) as the target for replacement. 
The results are shown in \cref{tab:ablation}. 
On the one hand, we can conclude that discriminating between the original caption and these altered ones is much easier when we randomly replace the words in the caption, even when only one word is changed. 
This margin is greater when we modify the nouns than when we modify the verbs in the captions, which aligns with our observation that current models rely heavily on static clues to make predictions. 
This demonstrates that the temporal understanding addressed by our VITATECS is more difficult to solve than simple object or action replacement.
Also, the accuracy of the model rises quickly as we increase the number of replaced words, while our VITATECS maintains its difficulty despite showing greater lingual diversity.
On the other hand, replacing words with their synonyms without contextual information may change their semantics significantly, as evidenced by the relatively high accuracy of models on these counterfactuals compared to VITATECS. 
This cautions us against the use of purely lexical methods for counterfactual construction. 
Finally, neither of these replacement methods is able to attach fine-grained labels to the resulting sentence, demonstrating the superiority of our counterfactual design. 


\section{Conclusion}

This work aims to address the deficiency of temporal understanding evaluation abilities in existing video-language datasets. 
We present a fine-grained characterization of temporal concepts in video descriptions, and introduce a novel dataset that measures the temporal understanding capabilities of VidLMs by their ability to distinguish between the actual description of a video and its temporally modified alternative. 
To facilitate dataset construction, we design a human-in-the-loop annotation framework by leveraging LLMs for counterfactual description generation. 
Evaluation of state-of-the-art models demonstrates their failure to fully grasp temporal concepts. 
We hope our work can provide valuable insight into the future development of video-language understanding research. 

\section*{Acknowledgement}
We thank all the anonymous reviewers for their constructive comments. This work is supported in part by a Huawei Research Grant, The National Natural Science Foundation of China (No. 62176002), and The Fundamental Research Funds for the Central Universities. Xu Sun is the corresponding author of this paper.

%
%

\begin{thebibliography}{10}
\providecommand{\url}[1]{\texttt{#1}}
\providecommand{\urlprefix}{URL }
\providecommand{\doi}[1]{https://doi.org/#1}

\bibitem{Bagad2023TestOT}
Bagad, P., Tapaswi, M., Snoek, C.G.M.: Test of time: Instilling video-language
  models with a sense of time. ArXiv  \textbf{abs/2301.02074} (2023)

\bibitem{Bain2022CLIPHitchhiker}
Bain, M., Nagrani, A., Varol, G., Zisserman, A.: A clip-hitchhiker's guide to
  long video retrieval. ArXiv  \textbf{abs/2205.08508} (2022)

\bibitem{speednet}
Benaim, S., Ephrat, A., Lang, O., Mosseri, I., Freeman, W.T., Rubinstein, M.,
  Irani, M., Dekel, T.: Speednet: Learning the speediness in videos. In:
  {CVPR}. pp. 9919--9928. Computer Vision Foundation / {IEEE} (2020)

\bibitem{gpt3}
Brown, T.B., Mann, B., Ryder, N., Subbiah, M., Kaplan, J., Dhariwal, P.,
  Neelakantan, A., Shyam, P., Sastry, G., Askell, A., Agarwal, S.,
  Herbert-Voss, A., Krueger, G., Henighan, T.J., Child, R., Ramesh, A.,
  Ziegler, D.M., Wu, J., Winter, C., Hesse, C., Chen, M., Sigler, E., Litwin,
  M., Gray, S., Chess, B., Clark, J., Berner, C., McCandlish, S., Radford, A.,
  Sutskever, I., Amodei, D.: Language models are few-shot learners. ArXiv
  \textbf{abs/2005.14165} (2020)

\bibitem{Buch2022RevisitingT}
Buch, S., Eyzaguirre, C., Gaidon, A., Wu, J., Fei-Fei, L., Niebles, J.C.:
  Revisiting the “video” in video-language understanding. 2022 IEEE/CVF
  Conference on Computer Vision and Pattern Recognition (CVPR) pp. 2907--2917
  (2022)

\bibitem{Chen2011MSVD}
Chen, D.L., Dolan, W.B.: Collecting highly parallel data for paraphrase
  evaluation. In: Annual Meeting of the Association for Computational
  Linguistics (2011)

\bibitem{cheng2022vindlu}
Cheng, F., Wang, X., Lei, J., Crandall, D., Bansal, M., Bertasius, G.: Vindlu:
  A recipe for effective video-and-language pretraining. arXiv preprint
  arXiv:2212.05051  (2022)

\bibitem{Choi2019WhyCI}
Choi, J., Gao, C., Messou, J.C., Huang, J.B.: Why can't i dance in the mall?
  learning to mitigate scene bias in action recognition. In: Neural Information
  Processing Systems (2019)

\bibitem{Diwan2022WhyIW}
Diwan, A., Berry, L., Choi, E., Harwath, D.F., Mahowald, K.: Why is winoground
  hard? investigating failures in visuolinguistic compositionality. In:
  Conference on Empirical Methods in Natural Language Processing (2022),
  \url{https://api.semanticscholar.org/CorpusID:253255481}

\bibitem{Dong2022ASF}
Dong, Q., Li, L., Dai, D., Zheng, C., Wu, Z., Chang, B., Sun, X., Xu, J., Sui,
  Z.: A survey for in-context learning. ArXiv  \textbf{abs/2301.00234} (2022)

\bibitem{evans200422}
Evans, V.: 22 how we conceptualise time: language, meaning and temporal
  cognition. The cognitive linguistics reader p.~733 (2004)

\bibitem{fu2021violet}
Fu, T.J., Li, L., Gan, Z., Lin, K., Wang, W.Y., Wang, L., Liu, Z.: Violet:
  End-to-end video-language transformers with masked visual-token modeling. In:
  arXiv:2111.1268 (2021)

\bibitem{Fu2022violetv2}
Fu, T.J., Li, L., Gan, Z., Lin, K., Wang, W.Y., Wang, L., Liu, Z.: An empirical
  study of end-to-end video-language transformers with masked visual modeling.
  ArXiv  \textbf{abs/2209.01540} (2022)

\bibitem{Gao2021CLIP2TVAE}
Gao, Z., Liu, J., Chen, S., Chang, D., Zhang, H., Yuan, J.: Clip2tv: An
  empirical study on transformer-based methods for video-text retrieval. ArXiv
  \textbf{abs/2111.05610} (2021)

\bibitem{Ghodrati2018VideoTP}
Ghodrati, A., Gavves, E., Snoek, C.G.M.: Video time: Properties, encoders and
  evaluation. In: British Machine Vision Conference (2018)

\bibitem{Girdhar2019CATERAD}
Girdhar, R., Ramanan, D.: Cater: A diagnostic dataset for compositional actions
  and temporal reasoning. ArXiv  \textbf{abs/1910.04744} (2019)

\bibitem{Gorti2022XPoolCL}
Gorti, S.K., Vouitsis, N., Ma, J., Golestan, K., Volkovs, M., Garg, A., Yu, G.:
  X-pool: Cross-modal language-video attention for text-video retrieval. 2022
  IEEE/CVF Conference on Computer Vision and Pattern Recognition (CVPR) pp.
  4996--5005 (2022)

\bibitem{Goyal2017TheS}
Goyal, R., Kahou, S.E., Michalski, V., Materzynska, J., Westphal, S., Kim, H.,
  Haenel, V., Fr{\"u}nd, I., Yianilos, P.N., Mueller-Freitag, M., Hoppe, F.,
  Thurau, C., Bax, I., Memisevic, R.: The “something something” video
  database for learning and evaluating visual common sense. 2017 IEEE
  International Conference on Computer Vision (ICCV) pp. 5843--5851 (2017)

\bibitem{Hendricks2018LocalizingMI}
Hendricks, L.A., Wang, O., Shechtman, E., Sivic, J., Darrell, T., Russell,
  B.C.: Localizing moments in video with temporal language. In: Conference on
  Empirical Methods in Natural Language Processing (2018)

\bibitem{Huang2018WhatMA}
Huang, D.A., Ramanathan, V., Mahajan, D.K., Torresani, L., Paluri, M., Fei-Fei,
  L., Niebles, J.C.: What makes a video a video: Analyzing temporal information
  in video understanding models and datasets. 2018 IEEE/CVF Conference on
  Computer Vision and Pattern Recognition pp. 7366--7375 (2018)

\bibitem{Huang2021SelfsupervisedVR}
Huang, L., Liu, Y., Wang, B., Pan, P., Xu, Y., Jin, R.: Self-supervised video
  representation learning by context and motion decoupling. 2021 IEEE/CVF
  Conference on Computer Vision and Pattern Recognition (CVPR) pp. 13881--13890
  (2021)

\bibitem{klein2013time}
Klein, W.: Time in language. routledge (2013)

\bibitem{Krishna2017DenseCaptioningEI}
Krishna, R., Hata, K., Ren, F., Fei-Fei, L., Niebles, J.C.: Dense-captioning
  events in videos. 2017 IEEE International Conference on Computer Vision
  (ICCV) pp. 706--715 (2017)

\bibitem{Lei2022RevealingSF}
Lei, J., Berg, T.L., Bansal, M.: Revealing single frame bias for
  video-and-language learning. ArXiv  \textbf{abs/2206.03428} (2022)

\bibitem{lei2021clipbert}
Lei, J., Li, L., Zhou, L., Gan, Z., Berg, T.L., Bansal, M., Liu, J.: Less is
  more: Clipbert for video-and-language learning via sparse sampling. In: CVPR
  (2021)

\bibitem{li2021alpro}
Li, D., Li, J., Li, H., Niebles, J.C., Hoi, S.C.H.: Align and prompt:
  Video-and-language pre-training with entity prompts. 2022 IEEE/CVF Conference
  on Computer Vision and Pattern Recognition (CVPR) pp. 4943--4953 (2021)

\bibitem{Li2022BLIPBL}
Li, J., Li, D., Xiong, C., Hoi, S.C.H.: Blip: Bootstrapping language-image
  pre-training for unified vision-language understanding and generation. In:
  International Conference on Machine Learning (2022)

\bibitem{Li2023VideoChatCV}
Li, K., He, Y., Wang, Y., Li, Y., Wang, W., Luo, P., Wang, Y., Wang, L., Qiao,
  Y.: Videochat: Chat-centric video understanding. ArXiv
  \textbf{abs/2305.06355} (2023),
  \url{https://api.semanticscholar.org/CorpusID:258588306}

\bibitem{Li2020HeroHE}
Li, L., Chen, Y.C., Cheng, Y., Gan, Z., Yu, L., Liu, J.: Hero: Hierarchical
  encoder for video+language omni-representation pre-training. In: Conference
  on Empirical Methods in Natural Language Processing (2020)

\bibitem{Li2022LAVENDERUV}
Li, L., Gan, Z., Lin, K., Lin, C.C., Liu, Z., Liu, C., Wang, L.: Lavender:
  Unifying video-language understanding as masked language modeling. ArXiv
  \textbf{abs/2206.07160} (2022)

\bibitem{diving48}
Li, Y., Li, Y., Vasconcelos, N.: {RESOUND:} towards action recognition without
  representation bias. In: {ECCV} {(6)}. Lecture Notes in Computer Science,
  vol. 11210, pp. 520--535. Springer (2018)

\bibitem{Liu2022WANLIWA}
Liu, A., Swayamdipta, S., Smith, N.A., Choi, Y.: Wanli: Worker and ai
  collaboration for natural language inference dataset creation. ArXiv
  \textbf{abs/2201.05955} (2022)

\bibitem{fetv}
Liu, Y., Li, L., Ren, S., Gao, R., Li, S., Chen, S., Sun, X., Hou, L.: {FETV:}
  {A} benchmark for fine-grained evaluation of open-domain text-to-video
  generation. In: NeurIPS (2023)

\bibitem{Luo2020UniViLMAU}
Luo, H., Ji, L., Shi, B., Huang, H., Duan, N., Li, T., Chen, X., Zhou, M.:
  Univilm: A unified video and language pre-training model for multimodal
  understanding and generation. ArXiv  \textbf{abs/2002.06353} (2020)

\bibitem{Luo2021CLIP4ClipAE}
Luo, H., Ji, L., Zhong, M., Chen, Y., Lei, W., Duan, N., Li, T.: Clip4clip: An
  empirical study of clip for end to end video clip retrieval. Neurocomputing
  \textbf{508},  293--304 (2021)

\bibitem{Ma2022XCLIPEM}
Ma, Y., Xu, G., Sun, X., Yan, M., Zhang, J.C., Ji, R.: X-clip: End-to-end
  multi-grained contrastive learning for video-text retrieval. Proceedings of
  the 30th ACM International Conference on Multimedia  (2022)

\bibitem{videochatgpt}
Maaz, M., Rasheed, H.A., Khan, S.H., Khan, F.S.: Video-chatgpt: Towards
  detailed video understanding via large vision and language models. CoRR
  \textbf{abs/2306.05424} (2023)

\bibitem{Meng2022GeneratingTD}
Meng, Y., Huang, J., Zhang, Y., Han, J.: Generating training data with language
  models: Towards zero-shot language understanding. ArXiv
  \textbf{abs/2202.04538} (2022)

\bibitem{chatgpt}
OpenAI: Introducing {ChatGPT} (2022), \url{https://openai.com/blog/chatgpt}

\bibitem{Park2022ExposingTL}
Park, J.S., Shen, S., Farhadi, A., Darrell, T., Choi, Y., Rohrbach, A.:
  Exposing the limits of video-text models through contrast sets. In: North
  American Chapter of the Association for Computational Linguistics (2022)

\bibitem{retro-actions}
Price, W., Damen, D.: Retro-actions: Learning 'close' by time-reversing 'open'
  videos. In: {ICCV} Workshops. pp. 1371--1380. {IEEE} (2019)

\bibitem{Radford2021CLIP}
Radford, A., Kim, J.W., Hallacy, C., Ramesh, A., Goh, G., Agarwal, S., Sastry,
  G., Askell, A., Mishkin, P., Clark, J., Krueger, G., Sutskever, I.: Learning
  transferable visual models from natural language supervision. In:
  International Conference on Machine Learning (2021)

\bibitem{reimers-2019-sentence-bert}
Reimers, N., Gurevych, I.: Sentence-bert: Sentence embeddings using siamese
  bert-networks. In: Proceedings of the 2019 Conference on Empirical Methods in
  Natural Language Processing. Association for Computational Linguistics (11
  2019), \url{http://arxiv.org/abs/1908.10084}

\bibitem{Ren2023TESTA}
Ren, S., Chen, S., Li, S., Sun, X., Hou, L.: {TESTA:} temporal-spatial token
  aggregation for long-form video-language understanding. In: {EMNLP}
  (Findings). pp. 932--947. Association for Computational Linguistics (2023)

\bibitem{timechat}
Ren, S., Yao, L., Li, S., Sun, X., Hou, L.: Timechat: {A} time-sensitive
  multimodal large language model for long video understanding. CoRR
  \textbf{abs/2312.02051} (2023)

\bibitem{Rohrbach2015LSMDC}
Rohrbach, A., Rohrbach, M., Tandon, N., Schiele, B.: A dataset for movie
  description. 2015 IEEE Conference on Computer Vision and Pattern Recognition
  (CVPR) pp. 3202--3212 (2015)

\bibitem{Schick2021GeneratingDW}
Schick, T., Sch{\"u}tze, H.: Generating datasets with pretrained language
  models. ArXiv  \textbf{abs/2104.07540} (2021)

\bibitem{SevillaLara2019OnlyTC}
Sevilla-Lara, L., Zha, S., Yan, Z., Goswami, V., Feiszli, M., Torresani, L.:
  Only time can tell: Discovering temporal data for temporal modeling. 2021
  IEEE Winter Conference on Applications of Computer Vision (WACV) pp. 535--544
  (2019)

\bibitem{finegym}
Shao, D., Zhao, Y., Dai, B., Lin, D.: Finegym: {A} hierarchical video dataset
  for fine-grained action understanding. In: {CVPR}. pp. 2613--2622. Computer
  Vision Foundation / {IEEE} (2020)

\bibitem{Thrush2022WinogroundPV}
Thrush, T., Jiang, R., Bartolo, M., Singh, A., Williams, A., Kiela, D., Ross,
  C.: Winoground: Probing vision and language models for visio-linguistic
  compositionality. 2022 IEEE/CVF Conference on Computer Vision and Pattern
  Recognition (CVPR) pp. 5228--5238 (2022)

\bibitem{Wang2022AllIO}
Wang, A., Ge, Y., Yan, R., Ge, Y., Lin, X., Cai, G., Wu, J., Shan, Y., Qie, X.,
  Shou, M.Z.: All in one: Exploring unified video-language pre-training. ArXiv
  \textbf{abs/2203.07303} (2022)

\bibitem{Wang2020RemovingTB}
Wang, J., Gao, Y., Li, K., Lin, Y., Ma, A.J., Sun, X.: Removing the background
  by adding the background: Towards background robust self-supervised video
  representation learning. 2021 IEEE/CVF Conference on Computer Vision and
  Pattern Recognition (CVPR) pp. 11799--11808 (2020)

\bibitem{Wang2022OmniVLOF}
Wang, J., Chen, D., Wu, Z., Luo, C., Zhou, L., Zhao, Y., Xie, Y., Liu, C.,
  Jiang, Y.G., Yuan, L.: Omnivl: One foundation model for image-language and
  video-language tasks. ArXiv  \textbf{abs/2209.07526} (2022)

\bibitem{Wang2022DisentangledRL}
Wang, Q., Zhang, Y., Zheng, Y., Pan, P., Hua, X.: Disentangled representation
  learning for text-video retrieval. ArXiv  \textbf{abs/2203.07111} (2022)

\bibitem{Wang2019VaTeXAL}
Wang, X.E., Wu, J., Chen, J., Li, L., fang Wang, Y., Wang, W.Y.: Vatex: A
  large-scale, high-quality multilingual dataset for video-and-language
  research. 2019 IEEE/CVF International Conference on Computer Vision (ICCV)
  pp. 4580--4590 (2019)

\bibitem{Wang2018PullingAO}
Wang, Y., Hoai, M.: Pulling actions out of context: Explicit separation for
  effective combination. 2018 IEEE/CVF Conference on Computer Vision and
  Pattern Recognition pp. 7044--7053 (2018)

\bibitem{DBLP:conf/cvpr/WeiLZF18}
Wei, D., Lim, J.J., Zisserman, A., Freeman, W.T.: Learning and using the arrow
  of time. In: {CVPR}. pp. 8052--8060. Computer Vision Foundation / {IEEE}
  Computer Society (2018)

\bibitem{star}
Wu, B., Yu, S., Chen, Z., Tenenbaum, J., Gan, C.: {STAR:} {A} benchmark for
  situated reasoning in real-world videos. In: NeurIPS Datasets and Benchmarks
  (2021)

\bibitem{Xiao2021NExTQANP}
Xiao, J., Shang, X., Yao, A., Chua, T.S.: Next-qa: Next phase of
  question-answering to explaining temporal actions. 2021 IEEE/CVF Conference
  on Computer Vision and Pattern Recognition (CVPR) pp. 9772--9781 (2021)

\bibitem{Xu2016MSRVTTAL}
Xu, J., Mei, T., Yao, T., Rui, Y.: Msr-vtt: A large video description dataset
  for bridging video and language. 2016 IEEE Conference on Computer Vision and
  Pattern Recognition (CVPR) pp. 5288--5296 (2016)

\bibitem{Xue2022CLIPViPAP}
Xue, H., Sun, Y., Liu, B., Fu, J., Song, R., Li, H., Luo, J.: Clip-vip:
  Adapting pre-trained image-text model to video-language representation
  alignment. ArXiv  \textbf{abs/2209.06430} (2022)

\bibitem{Ye2022ProGenPZ}
Ye, J., Gao, J., Feng, J., Wu, Z., Yu, T., Kong, L.: Progen: Progressive
  zero-shot dataset generation via in-context feedback. In: Conference on
  Empirical Methods in Natural Language Processing (2022)

\bibitem{Yi2019CLEVRERCE}
Yi, K., Gan, C., Li, Y., Kohli, P., Wu, J., Torralba, A., Tenenbaum, J.B.:
  Clevrer: Collision events for video representation and reasoning. ArXiv
  \textbf{abs/1910.01442} (2019)

\bibitem{Yu2019ActivityNetQAAD}
Yu, Z., Xu, D., Yu, J., Yu, T., Zhao, Z., Zhuang, Y., Tao, D.: Activitynet-qa:
  A dataset for understanding complex web videos via question answering. ArXiv
  \textbf{abs/1906.02467} (2019)

\bibitem{Yun2022TimeIM}
Yun, S., Kim, J., Han, D., Song, H., Ha, J.W., Shin, J.: Time is matter:
  Temporal self-supervision for video transformers. ArXiv
  \textbf{abs/2207.09067} (2022)

\bibitem{Zhang2023VideoLLaMAAI}
Zhang, H., Li, X., Bing, L.: Video-llama: An instruction-tuned audio-visual
  language model for video understanding. ArXiv  \textbf{abs/2306.02858}
  (2023), \url{https://api.semanticscholar.org/CorpusID:259075356}

\bibitem{Zhang2020DCADC}
Zhang, Z., Yin, Z., Ren, S., Li, X., Li, S.: Dca: Diversified co-attention
  towards informative live video commenting. In: Natural Language Processing
  and Chinese Computing (2020)

\bibitem{Zhou2017YouCook}
Zhou, L., Xu, C., Corso, J.J.: Towards automatic learning of procedures from
  web instructional videos. In: AAAI Conference on Artificial Intelligence
  (2017)

\end{thebibliography}

\end{document}